# Monocular 3D Object Detection using Multi-Stage Approaches with Attention and Slicing aided hyper inference

Ashish Patel,
Research Scientist,
ashishpatel.ce.2011@gmail.com

Abonia Sojasingarayar,
Machine Learning Scientist,
aboniaa@gmail.com

**Abstract**: *3D object detection is vital as it would enable us to capture objects' sizes, orientation, and position in the world. As a result, we would be able to use this 3D detection in real-world applications such as Augmented Reality (AR), self-driving cars, and robotics which perceive the world the same way we do as humans. Monocular 3D Object Detection is the task to draw 3D bounding box around objects in a single 2D RGB image. It is localization task but without any extra information like depth or other sensors or multiple images. Monocular 3D object detection is an important yet challenging task. Beyond the significant progress in image-based 2D object detection, 3D understanding of real-world objects is an open challenge that has not been explored extensively thus far. In addition to the most closely related studies.*

## I. Introduction

Three-dimensional (3D) object detection is a fundamental problem and enables various applications such as autonomous driving. Previous methods have achieved superior performance based on the accurate depth information from multiple sensors, such as LiDAR signal or stereo. To lower the sensor costs, some image-only monocular 3D object detection methods have been proposed and made impressive progress relying on geometry constraints between 2D and 3D. So, with this paper we will give an overview of different existing framework implemented for 3D detection and how those differ from one another. There are many benchmark datasets used for 3D object detection in this era which helps in improving the model performance.

The task of target detection is to find all Region of Interests (ROI) in the image and determine their positions and categories. Due to the different appearance, shape, and attitude of various objects, as well as the interference of lighting, shielding and other factors during imaging, object detection has continuously been a challenging problem in the field of computer vision. [1]

## II. Dataset

It is possible to use many benchmark dataset. we can come across few of the following: PASCAL3D+ [2] with RGB + 3D models data types of Indoor and contain more than 20,000 images. SUN RGB-D [3] with RGB-D data type of indoor and outdoor scene and contain more than 10,335 images with 800 categories. ObjectNet 3D [4] with RGB + 3D models data types of Indoor + Outdoor scenes and contains 90,127 images, FAT [5] with RGB + 3D models data types of Indoor + Outdoor scenes and contains 61,500 images, BOP [6] with RGB-D + 3D models data types of Indoor and contains more than 800 K train, test RGB-D (mostly synthetic) ,Objectron [6] with RGB data types of Indoor + Outdoor and contains more than 4 M (14,819 videos) images, KITTI 3D [7] with RGB (Stereo) + PointCloud data types of Driving Scenes and contains 14,999 images, CityScape 3D [8] with RGB (Stereo) data types of Driving Scenes and contains 5000 images, Synscapes[9] with RGB data types of Driving Scenes and contains 25,000 images, SYNTHIA-AL [10] with RGB data types of Driving Scenes and contains more than 143 K images.

**III. RELATED WORKS: Multistage Methods**

As monocular images lack depth information owing to the principle of perspective transformation, we can use deep learning to predict the depth map of the image first, which serves as the basis for 3D object detection in the next stage. To achieve effective monocular depth estimation, many algorithms have been developed in recent years. In addition to using the depth estimation module, the object ROI and depth feature map are fused to calculate the object coordinate and spatial location information. Using a multi-layer fusion scheme, this framework can generate the final pseudo-point cloud information for its application.

Likewise, it is also a popular algorithm for converting the image information into point cloud information; the point-cloud-related network is then used for processing. For another application of representation transform, the orthographic feature transform (OFT) maps perspective images to orthographic bird's eye view (BEV) images in the deep-learning-based framework. In general, the representation transform selects an application-specific data representation that is more suitable for the target scenario than the image domain. Hence, it can achieve satisfactory detection results.

*1. Depth Estimation*

On the basis of deep-learning-based monocular depth estimation, Xu and Chen [12] proposed the multi-level fusion-based 3D object detection (MF3D) algorithm, which combines the Deep3Dbox algorithm and a standard depth estimation module. Using deep CNN features, it basically uses the existing detectors. In addition to 2D proposals, the disparity estimation is computed to generate a 3D point cloud. Thus, the deep features derived from the RGB images and the point cloud are fused to enhance the object detection performance. The depth feature map and the ROI for objects are combined to obtain the 3D spatial location. The key idea of this framework is to use the multi-level fusion scheme, taking advantage of the standalone module for disparity computation. Experimental results showed that the performance of 3D object detection can be boosted by 3D localization. Below architecture shows sub-networks of the MF3D framework [12]. The task-specific modules are responsible for objectless classification, 2D box regression, and disparity prediction. Based on region proposals and point cloud maps from the estimated disparity, the 3D bounding box of the object is optimized and visualized as shown in the figures on the right.

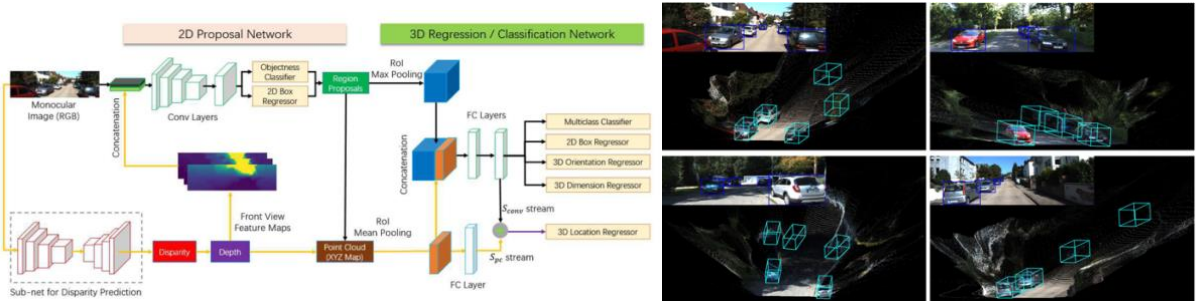

*Figure 1. Overview of MF3D for 3D object detection from monocular images.*
*Source : https://openaccess.thecvf.com/content_cvpr_2018/papers/Xu_Multi-Level_Fusion_Based_CVPR_2018_paper.pdf*

*2. FuDNN Approch and architecture*

A deep learning network based on PointRCNN, named FuDNN, is designed to realize 3D object detection. The architecture of FuDNN is shown in Figure 2, including a 2D backbone, a 3D backbone, an attention-based fusion sub-network, an RPN, and a 3D box refinement network. The 2D backbone is designed to learn 2D features from camera images automatically, and the attention-based fusion sub-network is proposed to fuse the 2D features with the 3D features extracted by PointNet++. The RPN and 3D box refinement network of PointRCNN are used to generate 3D proposals and refine the 3D.

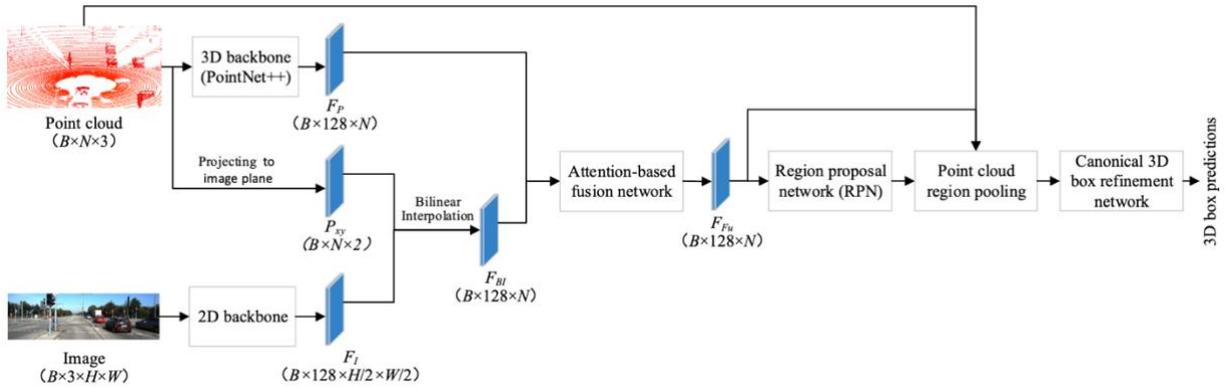

*Figure 2. Network architecture of the FuDNN. B, N, H, and W represent the batch size of FuDNN, the number of LiDAR points, the height, and width of image, respectively. Source : https://www.mdpi.com/2078-2489/13/4/169*

### 3. 3D shape information - Deep Manta approach

It is based on the Many-task CNN (Deep MANTA) which proposes accurate 2D vehicle bounding boxes using multiple refinement steps. The MANTA architecture also provides vehicle part coordinates (even if these parts are hidden), part visibility and 3D template for each detection. These fine features are then used to recover vehicle orientation and 3D localization using robust 2D/3D point matching. This approach outperforms state-of-the-art methods for vehicle detection and fine orientation estimation and clearly increases vehicle 3D localization compared to monocular approaches.

First, the input image is passed through the Deep MANTA network that outputs 2D scored bounding boxes, associated vehicle geometry (vehicle part coordinates, 3D template similarity) and part visibility properties. Instead of optimizing separate quantities, Chabot et al. [12] proposed a multi-tasking network structure for 2D and 3D vehicle analysis from a single image. For simultaneous part localization, visibility characterization, vehicle detection, and 3D dimension estimation, the many-tasks network (MANTA) first detects 2D bounding boxes of vehicles in multiple refinement stages. For each detection, it also gives the 3D shape template, part visibility, and part coordinates of the detected vehicle even if some parts are not visible. Then, these features are considered to estimate the vehicle localization and orientation using 2D–3D correspondence matching. To access the 3D information of the test objects, the vehicle models are searched for template matching. The real-time pose and orientation estimation uses the outputs of the network in the inference stage. At the time of publication [12], this approach was the state-of-the-art approach using the KITTI 3D benchmark in terms of vehicle detection, 3D localization, and orientation estimation tasks.

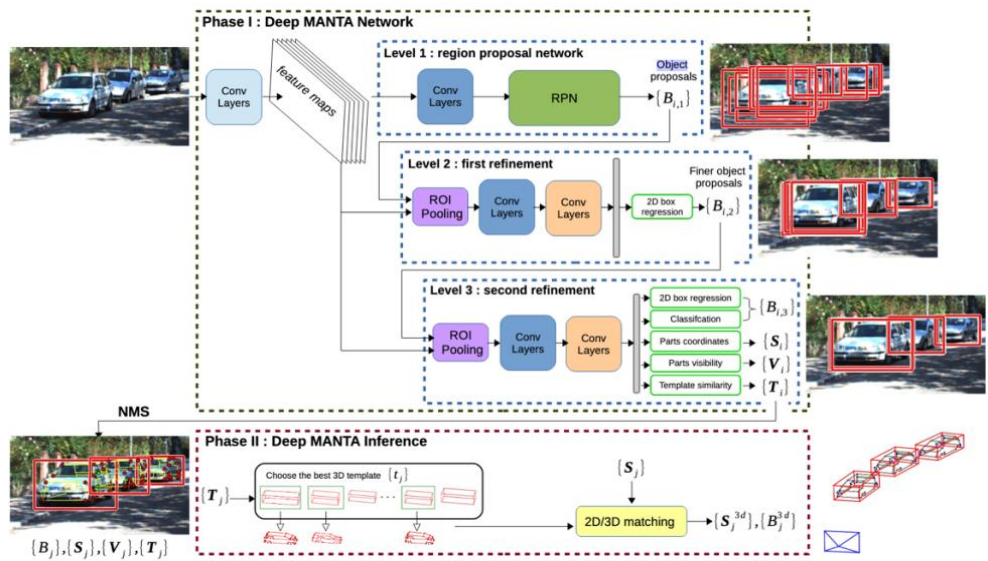

*Figure 3. Overview of the Deep MANTA approach. Source : https://arxiv.org/pdf/1703.07570.pdf*

# IV Our Proposal

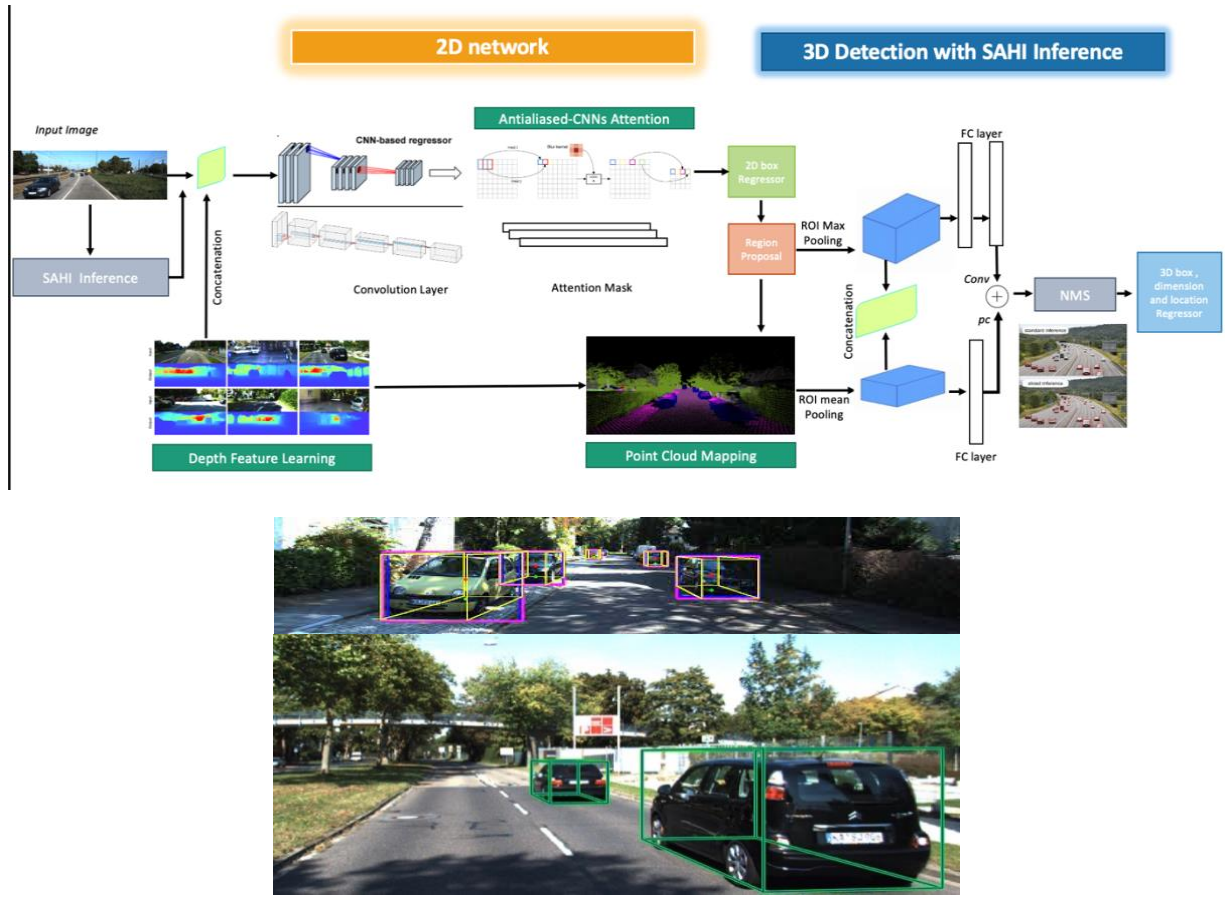

*Figure 4. Our Architecture for Monocular 3D Object Detection using Multi-Stage Approaches with Attention and Shift invariant Inference*

### 1. 2D Object detection and 3D monocular Object detection:

According to the detection procedure and mechanism, they can be categorizsd into two partss [15]. The first is one-stage detectors, they are applied over a dense sampling of possible object locations, such as OverFeat [18], YOLO [19, 20]. and SSD [16, 17], that can provide promising results with relatively fast speed. These methods trailed in accuracy even with a larger compute budge [15, 18]. The other is two-stage, proposal-driven detectors that apply classification and regression over learned features within object proposals. In the first stage, several methods are adopted for proposal generation. The widely used ones include BING [21], Selective Search [22], EdgeBoxes [23], DeepMask [24, 25] and RPN [26]. The most famous two-stage detector is the aforementioned Faster R-CNN, which can generate proposals and apply object recognition in an end-to-end fashion. Typically, two-stage detectors need fewer data augmentation tricks and have more accurate results in most public benchmarks .

2D object detectiontion and monocular 3D object detection are adopted on a single RGB picture. Unlike 2D object detection, it can be quite difficult for monocular 3D object detection since the lack of stereo information and accurate laser points from other sensors. Like two-stage, region-based 2D detectors, several works make use of high-quality 3D region proposals for further classification and detection. In [2], the author makes use of a general assumption that all the objects should lie close to the ground plane, which should be or fundamental to the image plane. The 3D object candidates are then exhaustively scored in the image plane by utilizing class segmentation, instance level segmentation, shape,contextual features and location priors. With the projected 2D proposals from 3D candidates, it uses Fast R-CNN to jointly predict category labels, bounding box offsets, and object orientation[12].

### 2. Depth Estimation:

*Depth estimation* is a crucial step towards inferring scene geometry from 2D images. The goal in *monocular depth estimation* is to predict the depth value of each pixel or inferring depth information, given only a single RGB image as input. It measures the distance of each pixel relative to the camera. Depth is extracted from either monocular (single) or stereo (multiple views of a scene) images. Traditional methods use multi-view geometry to find the relationship between the images. Newer methods can directly estimate depth by minimizing the regression loss, or by learning to generate a novel view from a sequence.

$$Disparity\ (D) = n1 - n2 = xl - xr$$

(Formula. 1)

$$Z = \frac{f}{d} * \frac{T}{D}$$

(Formula. 2)

Where, Z is Distance between point P and camera center, P is Target point in the physical world (scene point), PL is Point P in left camera image, PR is Point P in right camera image, n1 is Horizontal pixel distance of point PL in left camera image, n2 is Horizontal pixel distance of point PR in right camera image, T is Baseline distance between center of left and right cameras, f is Focal length of the camera d is Physical size of a pixel in camera sensor CMOS/CCD

### 3. Antialiased CNNs

Modern convolutional networks are not shift invariant, as small input shifts or translations can cause drastic changes in the output. Commonly used downsampling methods, such as max-pooling, strided-convolution, and average-pooling, ignore the sampling theorem. The well-known signal processing fix is anti-aliasing shown in Fig.5 by low pass filtering before downsampling. However, simply inserting this module into deep networks degrades performance; as a result, it is seldomly used today. We show that when integrated correctly, it is compatible with existing architectural components, such as max-pooling and strided-convolution.

Integrate classic anti-aliasing to improve shift-equivariance of deep networks. Critically, the method is compatible with existing downsampling strategies.
• Validate on common downsampling strategies – maxpooling, average-pooling, strided-convolution in different architectures. It can be test across multiple tasks – image classification and image-to-image translation.
• For ImageNet classification, surprisingly, that accuracy increases, indicating effective regularization.
• Furthermore, It can beobserve better generalization. Performance is more robust and stable to corruptions such as rotation, scaling, blurring, and noise variants[29].

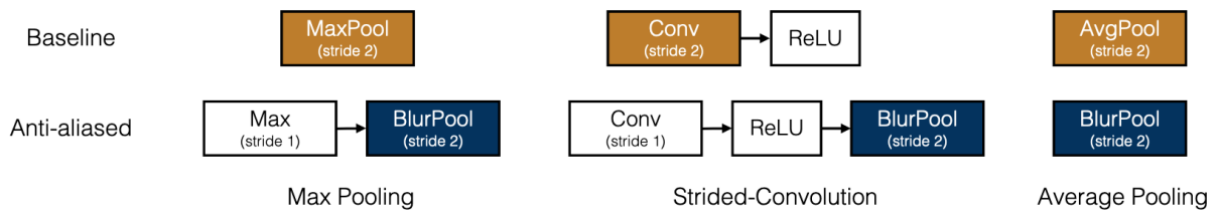

*Figure 5. Anti-aliasing common downsampling layers. (Top) Max-pooling, strided-convolution, and average-pooling can each be better antialiased (bottom) [29]*

### 4. Region Proposal

2D region proposal network, which can extract ROI from RGB images. Most recent frameworks are based on seminal work of Faster R-CNN due to the introduction of the region proposal network (RPN) as a highly effective method to efficiently generate object proposals. The RPN functions as a sliding window detector to check for the existence of objects at every spatial location of an image which match with a set of predefined template shapes, referred to as anchors. Despite that the RPN was conceived to be a preliminary stage within Faster R-CNN, it is often demonstrated to have promising effectiveness being extended to a single-shot standalone detector. We observe that 2D object detection performs reasonably and continues to make rapid advances. The 2D and 3D detection tasks each aim to ultimately classify all instances of an object; whereas they differ in the dimensionality of their localization targets. Intuitively, we expect the power of 2D detection can be leveraged to guide and improve the performance of 3D detection, ideally within a unified framework rather than as separate components. Hence, we propose to reformulate the 3D detection problem such that both 2D and 3D spaces utilize shared anchors and classification targets. In doing so, the 3D detecton is naturally able to perform on par with the performance of its 2D counterpart, from the perspective of reliably classifying objects. Therefore, the remaining challenge is reduced to 3D localization within the camera coordinate space. [30].

## 5. Point cloud Mapping

Point cloud mapping is a process in which a laser tracker locates several pre-designated points in a space and converts the information into a virtual map. Computing point cloud here means transforming the depth pixel from the depth image 2D coordinate system to the depth camera 3D coordinate system (*x, y* and *z*). The 3D coordinates are computed using the following formulas [2], where *depth (i, j)* is the depth value at the row *i* and column *j*:

$$\begin{cases} Z = depth(i,j) \\ x = (j - C_x) * Z / f_x \\ y = (i - C_y) * Z / f_y \end{cases}$$

(Formula. 3)

LiDAR point clouds, which provide the crucial 3D measurements,are indispensable in the labeling procedure. As a common practice, annotators annotate 3D boxes on the LiDAR points clouds. On the other hand, concerning the data collecting process in a self-driving system, a large number of successive snippets are collected. Generally speaking, to save the high annotation costs,only some key frames in collected snippets are labeled to train networks, such as KITTI dataset [7]. Consequently, massive LiDAR point clouds holding valuable 3D information remain unlabeled. Inspired by the 3D location label requirement and accurate LiDAR 3D measurements in 3D space, we propose a general and intuitive framework to make full use of LiDAR point clouds, dubbed LPCG (LiDAR point cloud guided monocular 3D object detection). Specifically, we use unlabeled LiDAR point clouds to generate pseudo labels, converting unlabeled data to training data for monocular 3D detectors. These pseudo labels are not as accurate as the manually annotated labels, but they are good enough for the training of monocular 3D detectors due to accurate LiDAR 3D measurements.LiDAR point clouds can provide valuable 3D location information. More specifically, LiDAR point clouds provide accurate depth measurement within the scene, which is crucial for 3D object detection as precise surrounding depths indicate locations of objects. Also, LiDAR point clouds can be easily captured by the LiDAR device, allowing a large amount of LiDAR point clouds to be collected offline without manual cost[31].

## 6. SAHI (Slicing Aided Hyper Inference)

In surveillance applications, detecting tiny items and objects that are far away in the scene is practically very difficult. Because such things are represented by a limited number of pixels in the image, traditional detectors have a tough time detecting them.Our proposal here is to use SAHI inference to leverage object detection. With SAHI, its possible to detect small objects. Slicing method is also utilized during the inference step as detailed in Fig. 6. First, the original query image I is sliced into l number of M × N overlapping patches $P^I_1, P^I_2 ... ... P^I_l$ . Then, each patch is resized while preserving the aspect ratio. After that, object detection forward pass is applied independently to each overlapping patch. An optional full-inference (FI) using the original image can be applied to detect larger objects. Finally, the overlapping prediction results and, if used, FI results are merged back into original size using NMS. During NMS, boxes having higher Intersection over Union (IoU) ratios than a predefined matching threshold Tm are matched and for each match, detections having detection probability than lower than Td are removed [33]

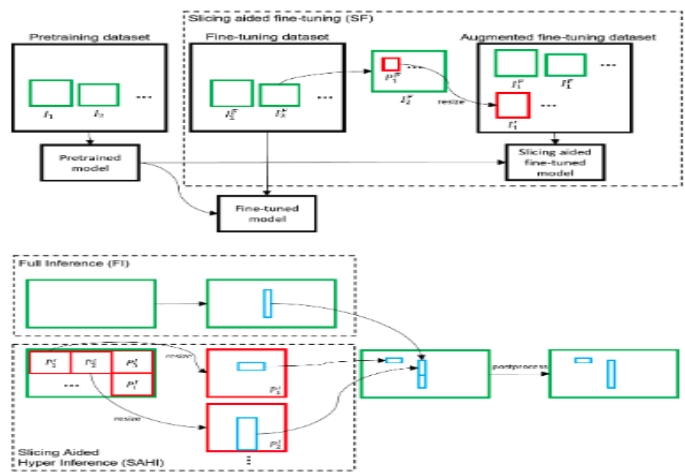

*Figure 6. Slicing aided fine-tuning (top) and slicing aided hyper inference (bottom) methods. In finetuning, the dataset is augmented by extracting patches from the images and resizing them to a larger size. During inference, image is divided into smaller patches and predictions are generated from larger resized versions of these patches. Then these predictions are converted back into original image coordinates after NMS.Optionally, predictions from full inference can also be added. [33]*

### 7. 3D Localization

It is much more complicated to estimate the 3D location of a 3D object. In our previous branches, only features inside region proposals are utilized for angle and dimension regression. However, estimating 3D location in the same way is difficult since the existence of RoI max pooling. Features generated from RoI max pooling have several drawbacks. First, it converts the features inside RoIs with different scales into fixed-sized feature tensors. Thus, it partly eliminates a fundamental photography constraint that the RoI with bigger size should lie closer to the camera. Besides, the image coordinates of each RoI are only used to fetch the corresponding region on the feature map.It means that different spatial locations of RoIs may have similar output after RoI max pooling, while the actual 3D location can differ a lot. In 2D bounding boxes regression,the absolute coordinates are achieved by estimating the offset to the proposals. However, for 3D localization, the 2D proposals do not contain 3D information of the coordinates [12]. To help localize the 3D object in our framework, the approximate layouts of objects or 3D locations of pixels can be modeled.

Disparity Estimation - To help understanding the whole scene in the image, a sub-net for disparity estimation from a single image is proposed.

Estimation Fusion for 3D Localization - With the estimated point cloud in the whole image as our prior knowledge and the previously generated RoIs, RoI mean pooling layer for 3D localization. The principle is quite simple and very similar to RoI max pooling.
Input Fusion with Front View Feature Maps Encoding - In addition to helping estimate the 3D locations, we also encode the estimated depth information with three-channel representations as the front view feature maps[12].

### V. Conclusion

In this Paper, we proposed a novel monocular 3D object detection training pipeline, which aims at simulating the feature learning behavior with strong attention-based detectors using Slicing aided hyper inference. We integrate low pass filtering to anti-alias, a common signal processing technique. The simple modification achieves higher consistency, across architectures and downsampling techniques slicing aided hyper inference scheme can directly be integrated into any object detection inference pipeline and does not require pretraining.Our Future work will be around implementing and comparing this architecture with methods used for state-of-art result.

# Acknowledgements